\documentclass{article}
\PassOptionsToPackage{numbers, sort&compress}{natbib}
\usepackage[preprint]{neurips_2022}

\usepackage[utf8]{inputenc} 
\usepackage[T1]{fontenc}    
\usepackage{hyperref}
\usepackage{url}            
\usepackage{booktabs}       
\usepackage{amsfonts}       
\usepackage[nice]{nicefrac}
\usepackage{microtype}      
\usepackage{xcolor}         

\usepackage{todonotes}
\usepackage[blackhypersetup]{tchdr}

\title{Conditional Permutation Invariant Flows}

\author{%
  Berend Zwartsenberg$^{1}$\thanks {Work done in part while at UBC. Email: \texttt{berend.zwartsenberg@inverted.ai}} \quad
  Adam \'Scibior$^{1}$ \quad
  Matthew Niedoba$^{1,2}$ \quad
  Vasileios Lioutas$^{1,2}$ \\
  \textbf{  Yunpeng Liu$^{1,2}$ \quad
  Justice Sefas$^{1,2}$ \quad   
  Setareh Dabiri$^{1}$ \quad
  Jonathan Wilder Lavington$^{1,2}$ }
 \\
  \textbf{
    Trevor Campbell$^{2}\thanks{Equal Contribution.}$ \qquad
  Frank Wood$^{1,2,3\dagger}$
  } \vspace{8pt}\\
  $^1$\normalfont{Inverted AI}\qquad $^2$University of British Columbia\qquad $^3$Mila
  }

\begin{document}
\maketitle

\begin{abstract}
We present a novel, conditional generative probabilistic model of set-valued data with a tractable log density.  This model is a continuous normalizing flow governed by permutation equivariant dynamics. These dynamics are driven by a learnable per-set-element term and pairwise interactions, both parametrized by deep neural networks.  We illustrate the utility of this model via applications including (1) complex traffic scene generation conditioned on visually specified map information, and (2) object bounding box generation conditioned directly on images.  We train our model by maximizing the expected likelihood of labeled conditional data under our flow, with the aid of a penalty that ensures the dynamics are smooth and hence efficiently solvable. Our method significantly outperforms non-permutation invariant baselines in terms of log likelihood and domain-specific metrics (offroad, collision, and combined infractions), yielding realistic samples that are difficult to distinguish from real data.

\end{abstract}

\section{Introduction}
Invariances built into neural network architectures can exploit symmetries to create more data efficient models. While these principles have long been known in discriminative modelling~\citep{Lecun1998,Cohen2015, Cohen2016, Finzi2021}, in particular permutation invariance has only recently become a topic of interest in generative models~\citep{Greff2019, Locatello2020}.
When learning a density that should be invariant to permutations we can either incorporate permutation invariance into the architecture of our deep generative model or we can factorially augment our observations and hope that the generative model architecture is sufficiently flexible to at least approximately learn a distribution that assigns the same mass to known equivalents. The former is vastly more data efficient but places restrictions on the kinds of architectures that can be utilized, which might lead one to worry about performance limitations.
While the latter does allow unrestricted architectures it is often is so data-inefficient that, despite the advantage of fewer limitations, achieving good performance is extremely challenging, to the point of being impossible.

In this work we describe a new approach to permutation invariant \emph{conditional} density estimation that, while architecturally restricted to achieve invariance,
is demonstrably flexible enough to achieve high performance on a number of non-trivial density estimation tasks.

\begin{figure*}[t]
  \centering
  \includegraphics[width=5.5in]{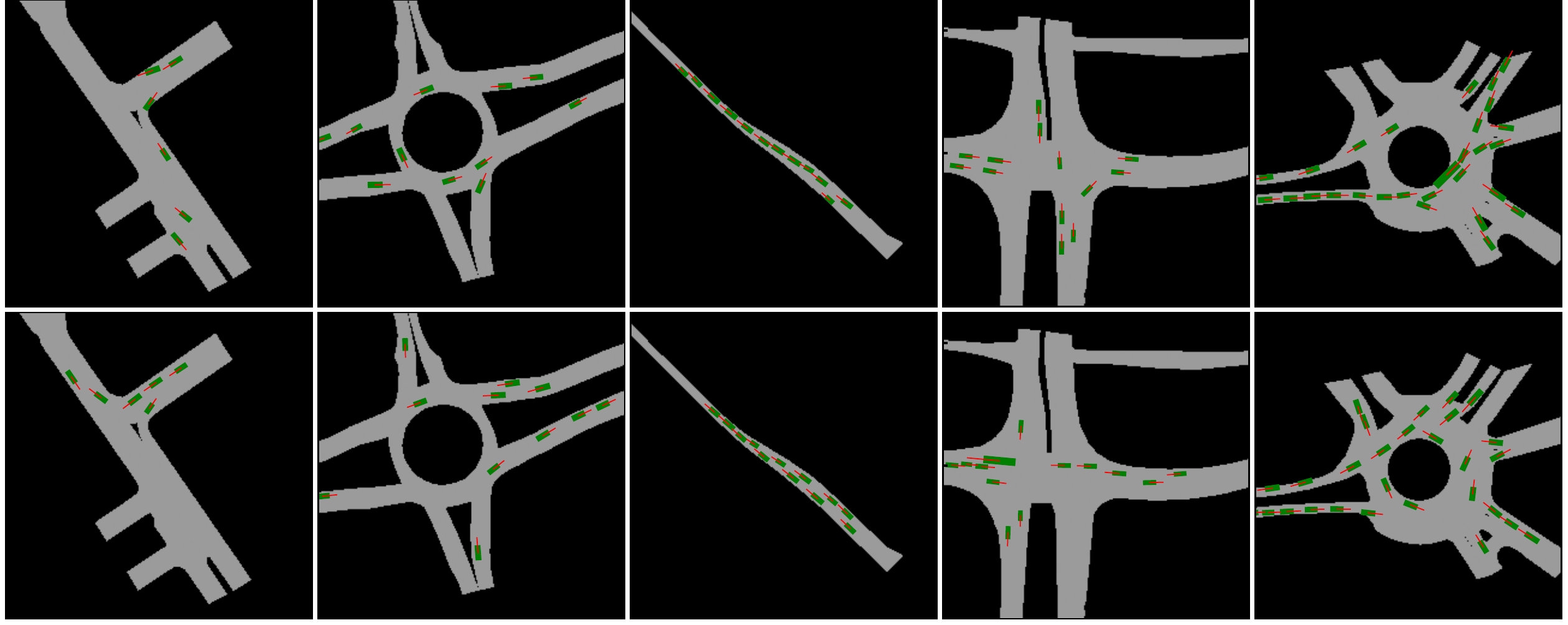}
\caption[]{\label{fig:initial_conditions_samples} Realistic vehicle placement as a permutation invariant modeling problem.  At every moment in time vehicles in the real world exhibit a characteristic spatial distribution of position, orientation, and size; notably vehicles (green rectangles) do not overlap, usually are correctly oriented (red lines indicate forward direction), and almost exclusively are conditionally distributed so as to be present only in driving lanes (shown in grey).  The likelihood of each such arrangement does not depend on the ordering of the vehicles (permutation invariance).  Each column shows a particular map with vehicle positions from real training data and from infraction free samples drawn from our permutation invariant flow conditioned on the map image.  Note that because the image indicates lanes, not drivable area, the training data includes examples of vehicles that hang over into the black.  We invite the reader to guess which image in each column is real and which is generated by our model.  The answer appears in a footnote at the end of the paper.\footnotemark{}
} 
\end{figure*}

Permutation invariant distributions, where the
likelihood of a collection of objects does not change if they are re-ordered, appear widely. The joint distribution of independent and identically distributed observations is permutation invariant, while in more complex examples the observations are no longer independent, but still exchangeable. 
Practical examples include the distribution of non-overlapping physical object locations in a scene, the set of potentially overlapping object bounding boxes given an image, and so forth (see \cref{fig:initial_conditions_samples}).  In all of these we know that the probability assigned to a set of such objects (i.e. locations, bounding boxes) should be invariant to the order of the objects in the joint distribution function argument list.

Recent work has addressed this problem by introducing equivariant
normalizing flows~\cite{Kohler2020, Satorras2021, Bilos2021}.  Our work builds on theirs but differs in subtle but key ways that increase the flexibility of our models.  More substantially this body of prior art addresses \textit{non-conditional} density estimation rather than \textit{conditional} density estimation as we do here.  
We believe that we are the first to develop \textit{conditional} permutation invariant flows.

We demonstrate our conditional permutation invariant flow on two difficult conditional density estimation tasks: realistic traffic scene generation (\cref{fig:initial_conditions_samples}) given a map and bounding box prediction given an image.  In both the set of permutation invariant objects is a set of oriented bounding boxes with additional associated semantic information such as heading.  We show that our method significantly outperforms baselines and meaningful ablations of our model.

\subsection{Background}

\subsubsection{Normalizing flows}
Normalizing Flows~\cite{Tabak2010, Tabak2013, Rezende2015} are probability distributions
that are constructed by combining a simple base distribution
$p_{\mathbf{z}}(\mathbf{z})$ (e.g., a standard normal) and a differentiable
transformation $T$ with differentiable inverse, that maps $\mathbf{z}$ to a
variable $\mathbf{x}$
\[\label{eqn:norm_flow_transformation}
  \mathbf{x} = T^{-1}(\mathbf{z}).
\]
We can then express the density using the change of variables
formula
\[\label{eqn:norm_flow_likelihood}
  p_{\mathbf{x}}(\mathbf{x}) = p_{\mathbf{z}}(T(\mathbf{x})) \left| \det \D{T^{-1}(\mathbf{z})}{\mathbf{z}} \Big|_{\mathbf{z} = T(\mathbf{x})} \right|^{-1},
\] where $p$ denotes the respective densities over variables $\mathbf{x}$ and $\mathbf{z}$ connected by transformation $T$ with inverse $T^{-1}$.
The transformation $T$ can be parametrized and used to approximate some distribution over data $\mathbf{x} \sim \pi$ by maximizing the likelihood of this data under the approximate distribution using gradient
descent. An important feature distinguishing normalizing flows from other models
is that in addition to a method to generate samples they provide a tractable log density, enabling maximum likelihood training and outlier detection among others. This
formulation, while powerful, has two noteworthy design challenges: the right hand side of \cref{eqn:norm_flow_likelihood} has to be efficiently evaluable and the aforementioned requirement that $T$ be invertible. The approach in the field generally is to define a chain of transformations $T_0 \circ \dots \circ T_n$, each of which satisfy both conditions. In this manner, they can be comparatively simple, yet when joined together provide a flexible approximating family.
\subsubsection{Continuous normalizing flows}
Continuous normalizing flows were first introduced in~\cite{Chen2018}, and then
further developed in~\cite{Grathwohl2018}. The concept is to use a
continuous transformation of variables, described by dynamics function $\mathbf{v}$ parametrized by a variable $t$ in the
form of an ordinary differential equation (ODE)
\[\label{eqn:continuous_var_transform}
  \mathbf{x}(t_{1}) = \mathbf{x}(t_{0}) + \int_{t_{0}}^{t_{1}}\mathbf{v}_{\theta}(\mathbf{x}(t), t) dt.
\] 
We set $\mathbf{x}(t_0) = \mathbf{z}$ and $\mathbf{x}(t_1) = \mathbf{x}$, so that \cref{eqn:continuous_var_transform} provides our definition of $T^{-1}$ as defined in \cref{eqn:norm_flow_transformation}. Similarly, if we integrate backward in time from $t_1$ to $t_0$ we obtain $T$. The dynamics $\mathbf{v}_{\theta}(\mathbf{x(t)}, t)$ can be represented by a flexible function. As long as the dynamics
function is uniformly Lipschitz continuous in $\mathbf{x}$ and uniformly
continuous in $t$, the solution to the ODE is unique, and the transformation is invertible~\cite{Coddington1955}. In this case, we can write the probability density as another
ODE~\cite{Grathwohl2018}
\[\label{eqn:log_prob_diff_eqn}
  \der{\log p_t \left(\mathbf{x}\left(t\right)\right)}{t} = - \nabla_{\mathbf{x}} \cdot \mathbf{v}_{\theta}(\mathbf{x}(t), t).
\]
The term on the right hand side is the \textit{divergence} (not gradient) of the dynamics (sometimes
equivalently written as the trace of the Jacobian,
note that $\mathbf{v}_{\theta}$ is vector valued in \cref{eqn:log_prob_diff_eqn}). Integrating this ODE from
the probability density at $t_{0}$ gives the density at $t_{1}$
\[\label{eqn:log_prob_transformation}
  \log p_{t_{1}}(\mathbf{x}(t_{1})) = \log p_{t_{0}}(\mathbf{x}(t_{0})) - \int_{t_{0}}^{t_{1}} \nabla_{\mathbf{x}} \cdot \mathbf{v}_{\theta}(\mathbf{x}(t), t) dt.
\]
\cref{eqn:log_prob_transformation} is the equivalent of
\cref{eqn:norm_flow_likelihood} for continuous normalizing flows. Together with a
suitable base distribution (e.g. a standard normal), the above
transformation constitutes a distribution with a tractable likelihood and
generative mechanism, which we will exploit to construct our flows.

\subsubsection{Invariance and equivariance}\label{ssec:invariance}
We seek to construct distributions that have a 
permutation invariant density via permutation equivariant
transformations. We state here the
definition of permutation invariance and equivariance we adopt.
\begin{ndefn}\label{dfn:permutation_definition}
Let $\mathbf{x} = \left(\mathbf{x}_1 \dots \mathbf{x}_N\right)$ where each $\mathbf{x}_n \in \reals^{D}$,  and
let permutations $\sigma$ act on $\mathbf{x}$ via
\[\label{eqn:permutation_operator}
  \sigma \mathbf{x} =  \left(\mathbf{x}_{\sigma_1} \dots \mathbf{x}_{\sigma_N}\right).
\]
A
  function $G\st \reals^{N \times D} \to \reals$ is permutation \emph{invariant} if for any permutation $\sigma$,
  \[
   \forall\, \mathbf{x} \in \reals^{N \times D}, \quad G(\sigma \mathbf{x}) = G(\mathbf{x}).
  \]
  A function $F \st \reals^{N \times D} \to \reals^{N \times D}$ is permutation \emph{equivariant} if for any permutation $\sigma$,
  \[
    \forall\, \mathbf{x} \in \reals^{N \times D}, \quad  F(\sigma \mathbf{x}) = \sigma F(\mathbf{x}).
  \]
\end{ndefn}

\subsection{Related work}
Permutation invariant models have been studied in the literature for some time.
Examples include models of sets~\cite{Zaheer2017, Lee2019}, and graphs~\cite{Duvenaud2015,
  Kipf2017, Kipf2018}. Recently, also generative models for sets have made an
appearance,~\cite{Zhang2019, Burgess2019, Greff2019, Locatello2020,
  Zhang2020}. 
Our conditional permutation invariant flows belong to the larger class of generative models, such as variational autoencoders~\cite{Kingma2013}, generative adversarial networks~\cite{Goodfellow2014}, and
normalizing flows \cite{Rezende2015}. Among these, normalizing flows are the only class of models that enables likelihood evaluation.

Other work belonging to the generative category is ``Equivariant Hamiltonian
flows'' \cite{Rezende2019}, which relates to our work since it models interactions elements of a
set using Hamiltonian dynamics. The choice of these dynamics allows
the use of a symplectic integrator, and the transformation is volume conserving,
eliminating the need to integrate a divergence term. However, this also
requires the introduction of a set of latent momentum variables that preclude the exact calculation of a density.

Our work is strongly related to, and draws inspiration from recent work that
uses continuous normalizing flows with permutation invariant dynamics~\cite{Kohler2020, Satorras2021, Bilos2021}. However~\cite{Kohler2020} and
\cite{Satorras2021} focus also on rotation and translation invariance, in
order to model molecular graphs. Out of this work,~\cite{Bilos2021} is closest to
ours, as it focuses on sets rather than equivariant graphs. But because it employs a dynamics function that focuses on being cheap to evaluate, it is less flexible as it ignores interaction terms. Importantly, none of this previous work considers the problem of
\emph{conditioning} on inputs and learning a distribution that is able to deal with a varying conditional input distribution.

\section{Conditional permutation invariant flows}
\subsection{Permutation invariant flows}
In this work, we will construct normalizing flows that are characterized by a permutation equivariant transformation
$T(\sigma \mathbf{x}) = \sigma T(\mathbf{x})$; we will demonstrate these flows produce a
permutation invariant density $p(\mathbf{x}) = p(\sigma \mathbf{x})$.
We construct our permutation invariant flows using a dynamics function that is
based on a global force term and pairwise interaction terms
\[\label{eqn:pairwise_dynamics}
  \mathbf{v}_{\theta,i}(\mathbf{x}) = \sum_{j \setminus i} f_{\theta}(\mathbf{x}_{i},\mathbf{x}_{j}) + g_{\theta}(\mathbf{x}_{i}).
\]
Here, $\mathbf{v}_{\theta,i} \in \reals^D$ denotes the $i^\text{th}$ element of $\mathbf{v}_{\theta}$, $\mathbf{x} \in \reals^{N \times D}$,
$g_{\theta} \st \reals^{D} \to \reals^{D}$, and
$f_{\theta} \st \reals^{2D} \to \reals^{D}$. This construction can be interpreted
as objects $\mathbf{x}_{i}$ moving in a global potential with
corresponding force field $g_{\theta}(\mathbf{x}_{i})$, and interacting with other
objects through the pairwise interaction $f_{\theta}(\mathbf{x}_{i}, \mathbf{x}_{j})$.
We proceed to construct a continuous normalizing flow using the function in
\cref{eqn:pairwise_dynamics} as the dynamics. If we use a permutation invariant
base distribution $p(\mathbf{x}(t_{0})) = p(\mathbf{z})$ we obtain the following:
\begin{nthm}\label{thm:perm_invariant}
  Let the
  transformation $\mathbf{z} = T(\mathbf{x})$ be as defined in
  \cref{eqn:continuous_var_transform}, with dynamics
  $\mathbf{v}_{\theta}(\mathbf{x})$ as defined in \cref{eqn:pairwise_dynamics}.
  If $p(\mathbf{z})$ is
  permutation invariant, then
 the transformation $T$ is permutation equivariant,
  and the density $p(\mathbf{x})$ is permutation invariant.
\end{nthm}
The proof of \cref{thm:perm_invariant} is given in \cref{sup:proof}. The
dynamical system is similar to a (classically) interacting set of particles
in a global potential. The dynamics as presented in
\cref{eqn:pairwise_dynamics} are independent of time; in a few cases, however, we
have found it useful to make the dynamics time-dependent, i.e.
$\mathbf{v}_{\theta}(\mathbf{x}, t)$, by passing time to both $g_{\theta}$ and $f_{\theta}$ as an input. A complete overview of when time dependence is used is given in the supplementary information \cref{sup:exp_params}. In practice we represent $f_{\theta}$ and $g_{\theta}$ by neural networks, which satisfy the criterion of uniform Lipschitz continuity if activation functions are chosen appropriately, thereby guaranteeing invertibility. More details on implementation can be found in \cref{sup:force_functions}.

\subsection{Divergence}
Given the dynamics in \cref{eqn:pairwise_dynamics}, we compute the density at time $t$ in \cref{eqn:log_prob_diff_eqn} using the divergence
\[\label{eqn:divergence}
  \nabla_{\mathbf{x}} \cdot \mathbf{v}_{\theta}(\mathbf{x}) =& \sum_{i,j,i\neq j} \nabla_{\mathbf{x}_{i}}\cdot f_{\theta}(\mathbf{x}_{i}, \mathbf{x}_{j})
  + \sum_{i} \nabla_{\mathbf{x}_{i}}\cdot g_{\theta}(\mathbf{x}_{i}).
\]
A na{\"i}ve computation of the divergence in \cref{eqn:log_prob_diff_eqn} using automatic differentiation is expensive, as computing the
Jacobian requires $ND$ evaluations, one for each of the $ND$ terms in  $\mathbf{v}_{\theta}$ \cite{Chen2018, Grathwohl2018}. Since the cost of
evaluating \cref{eqn:pairwise_dynamics} is quadratic in $N$, this would result in an asymptotic computational cost
of $N^{3}D^{2}$ for the forward and backward pass. Earlier work has suggested the use of the Hutchinson's trace
estimator \cite{Chen2018} for the divergence, which reduces the cost of the divergence to that of  $\mathbf{v}$, but suffers from high variance \cite{Chen2019}. Instead we opt to 
re-express the divergence of $\mathbf{v}_{\theta}$ in terms of derivatives of
$f(\mathbf{x}_{i}, \mathbf{x_{j}})$ and $g(\mathbf{x}_{i})$, resulting in \cref{eqn:divergence}. 
The form in \cref{eqn:divergence} is quadratic in $N$,
and therefore same cost in $N$ as the evaluation of $\mathbf{v}_{\theta}$ itself, both in the
forward and backward pass.

\subsection{Regularization}

Continuous normalizing flows have no inherent mechanism that penalizes very
complex dynamics. While in theory there is no reason to prefer simple dynamics,
in practice, the numerical integration can result in long computation times and sometimes numerical instabilities when using an
adaptive scheme. This effect has been previously observed in the literature, and
suggestions to regularize the dynamics have been proposed in previous
work~\cite{Finlay2020, Kelly2020}. While the work of~\cite{Kelly2020} is more
comprehensive, we find that an adaptation of the simpler solution proposed in~\cite{Finlay2020}
works well for our purposes. The proposed solution in~\cite{Finlay2020} is to
add a term proportional to the squared Frobenius norm of the Jacobian, and the
$\ell^{2}$-norm of the dynamics. We use the $\ell^{2}$-norm for the
dynamics; however, since we do not calculate or estimate the full Jacobian we calculate
\[ \label{eqn:regularization}
  \ell_{div}^{2} = \sum_{i\neq j, d,d^{\prime}}{\left(\D{f_{d}({x_{id}},{x_{jd}})}{x_{id^{\prime}}}\right)}^{2}
  + \sum_{i, d,d^{\prime}}{\left(\D{g_{d}({x_{id}})}{x_{id^{\prime}}}\right)}^{2}.
\]
We find that this penalty significantly reduces
the number of evaluations of our trained flows. We visualize the effect this penalty has on the dynamics in some examples in \cref{sup:regularization}.

\subsection{Conditional permutation invariant flows}
When learning a distribution artifact, making this object
dependent on an external input, and obtaining a flow that is valid for an entire distribution of such inputs is an essential tool to perform amortized inference, in which a family of posterior distributions is learned, given a distribution of observations.
An example would be to produce a valid distribution of a set of bounding
box locations and sizes $\mathbf{x}$ for objects in an image $\mathbf{y}$ selected from a distribution of images. 
We will denote the conditioning input as
$\mathbf{y}$, coming from some data distribution $\pi(\mathbf{y})$. To model such cases we construct a dynamics function that depends on $\mathbf{y}$ by modifying the
pair forces, and the global force to
$f_{\theta}(\mathbf{x}_{i},\mathbf{x}_{j},\mathbf{y})$ and
$g_{\theta}(\mathbf{x}_{i},\mathbf{y})$. The dynamics then become:
\[\label{eqn:conditional_dynamics}
  \mathbf{v}_{\theta,i}(\mathbf{x},\mathbf{y}) = \sum_{j \setminus i} f_{\theta}(\mathbf{x}_{i},\mathbf{x}_{j},\mathbf{y}) + g_{\theta}(\mathbf{x}_{i},\mathbf{y}).
\]
Note that here too, the dynamics can be made time dependent by passing time $t$ as an argument.

We will train our flows by minimizing the
Kullback-Leibler divergence to the joint distribution $\pi(\mathbf{x},\mathbf{y}) = \pi(\mathbf{x}|\mathbf{y})\pi(\mathbf{y})$ over data $\mathbf{x}$ and condition $\mathbf{y}$:
\newcommand{\EEu}{\mathop\EE}
\[
  \argmin_{\theta} \kl{\pi(\mathbf{x},\mathbf{y})}{p_\theta(\mathbf{x}|\mathbf{y})\pi(\mathbf{y})} = \argmin_{\theta}\EEu_{\mathbf{y} \sim \pi(\mathbf{y})}\kl{\pi(\mathbf{x}|\mathbf{y})}{p_\theta(\mathbf{x}|\mathbf{y}))}.
\]
In other words, we optimize our flow to match the distribution of $\mathbf{x}$ in \emph{expectation} over $\pi(\mathbf{y})$.

\section{Experiments}\label{sec:experiments}
\begin{figure*}[t]
  \centering
  \includegraphics[width=4.5in]{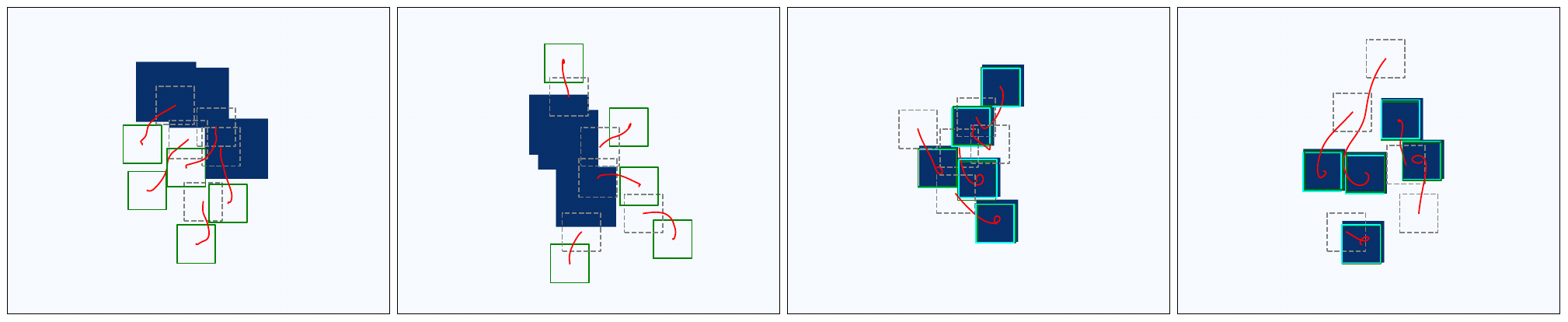}
\caption{\label{fig:toy_examples}
Two pedagogical permutation invariant modeling tasks. The left two panels illustrate the first task; \emph{conditionally} modeling non-overlapping squares (green), which also do not overlap with the blue boxes whose arrangement varies between datapoints. The right two panels illustrate the second task; modeling boxes that are conditionally distributed so as to exactly surround the underlying blue boxes. Samples from the base $p(\mathbf{z})$ and final distribution $p_{\theta}(\mathbf{x} | \mathbf{y})$ are plotted in
dashed grey and green lines respectively. The conditional input is
plotted as a blue on white image. Red lines indicate the trajectories the
objects follow by integrating the dynamics function $\mathbf{v}(\mathbf{x}(t), \mathbf{y})$.
}
\end{figure*}

\subsection{Synthetic examples}\label{sec:synthetic}

We start our experiments with  two pedagogical examples that demonstrate the capabilities and mechanisms of our
flows. These experiments roughly correspond to collision-avoidant object placement with environmental constraints, and bounding box prediction.  These examples show that the learned dynamics can  exhibit complex interactions that can be any combination of repulsive, attractive, or coordinating.  \cref{fig:toy_examples} shows results related to all three examples. 

The first
example task is to model the spatial distribution of five \emph{non-overlapping}  squares of width $w = 1$, that furthermore do not overlap with the prohibited regions shown in blue. This example is representative of placing assets into a physically realizable configuration in accordance with constraints imposed by an environment. We fit our conditional flow to a dataset generated by first sampling a prohibited region---three boxes of width $w = 1.5$ from a standard normal prior---and then sampling box locations independently from a standard normal prior, with rejection for overlap with previous boxes or the prohibited region, until a total of five boxes are sampled. The prohibited region is input to our conditional flow as an image tensor.  Since the dataset was generated via rejection sampling, we can compare our sample efficiency against it. The conditional flow provides a substantial sampling efficiency improvement  (77\% valid) over rejection sampling (0.02\% valid), in addition to a tractable density.

The second example task is bounding box prediction, or
conditionally generating object bounding boxes $\mathbf{x}$ directly from an image $\mathbf{y}$. Here the objects are monochrome blue squares. Data is generated in a similar manner as in the first experiment: squares are sampled indepenently from a standard normal prior, and rejected if they overlap. The conditional input is an image of the generated boxes. Sampled bounding boxes from our trained flow flow achieve an average intersection over union (IOU) of 0.85 with the ground truth bounding boxes. 

We display representative
samples and their trajectories through time in \cref{fig:toy_examples} for both experiments. In the left two panels it can be seen that initial
samples are transported around the space to avoid one another; in the right two, the boxes coordinate through the pairwise interactions to each surround exactly one of the objects in the scene.
Further details for these experiments appear in \cref{ssec:addtional_experimental_results}.

\subsection{Traffic scenes}\label{ssec:experiments_traffic}
Modeling and being able to sample realistic traffic scenes is an essential task related to autonomous driving simulation and control. Referring back to \cref{fig:initial_conditions_samples}, the problem---similar to the first pedagogical task above---is one of modeling the physical configuration a collection of agents conditioned on a
representation of the environment. Until recently, the predominant
methods for generating realistic vehicle configurations were rule-based~\cite{Yang1996, Lopez2018}. Rule-based systems can be tailored to have
 desirable properties such as avoiding occurences of offroad and vehicle overlap, but they produce vehicle arrangements that are distributionally dissimilar to real data.
Furthermore, depending on the particular design, such systems may
not have a tractable density, which limits their downstream utility and precludes quantification of the aforementioned sim-to-real gap. 
 
Recent work addressing this problem uses non-rule-based autoregressive model \cite{Tan2021} that enables sequential generation of vehicle and agent positions conditioned on a visual representation of a map. 
While this model-based approach closes the gap between simulation and reality, modeling sets autoregressively introduces
a factorial data augmentation requirement,
as there is no intrinsic ordering of actors. The
authors of~\cite{Tan2021} avoid this by imposing an arbitrary order, sampling agents from left to right. 
Our experiments indicate that, at least for this specific task, directly addressing permutation invariance is preferred, and avoids the need to arbitrarily fix the order of elements.

To test the performance of our flows on this task, we train them to generate a
scene of cars in the INTERACTION dataset~\cite{interactiondataset}, conditioned on a rendered image of the drivable area $\mathbf{y}$. The properties that our flows predict are
two-dimensional position, size, aspect ratio and orientation for each of $N$ agents,
i.e. $\mathbf{x} \in \reals^{N \times 5}$. An advantage of the formulation of
the dynamics in \cref{eqn:conditional_dynamics} is that they can be applied with
the same $f$ and $g$ regardless of the number of agents $N$. We make use of this
property, and train a single model on a varying
amount of agents, amortizing over the number of agents $N$.  At test time, generating $N$ agents is accomplished simply by initializing the flow appropriately.

We train our flows until the likelihood of the data stops increasing, or the likelihood of
a held out validation sets starts decreasing, whichever comes fist.
Examples of the data we train on, and representative samples from our trained flow are shown in \cref{fig:initial_conditions_samples}.

\subsubsection{Baselines}
We compare our conditional flows against several
baselines. Three are
non-permutation invariant flows: a unimodal Gaussian model, a RealNVP based
model~\cite{Dinh2017}, and a ``vanilla'' continuous normalizing flow (``CNF''). We also implement
an autoregressive model consisting of a convolutional neural net paired
with a recurrent neural network and a 10 component Gaussian mixture for every
prediction component, and adopt the canonical ordering discussed in
their paper. 
We also test two ablations of our model: one where the dynamics are restricted only to single particle terms $g_{\theta}$ (``PIF Single''), and one the dynamics only include the pair term $f_{\theta}$ (``PIF Pair''). 

To compare to non-permutation invariant methods, we have to fix the number of
agents, as such architectures cannot straightforwardly be provisioned at test time to generate and score sets that differ in cardinality to the training set. We exclude all data with less
than seven agents, and cases with more than seven agents are pruned to retain
only the agents closest to the center. The cardinality of seven was chosen to
retain as much data as possible while not making each individual scene too small. Furthermore, we restricted the INTERACTION dataset to the roundabout scenes in order to better match the 7 agent target while still maintaining a semantically similar set of possible $\mathbf{y}$. We compare the
negative log likelihood (NLL) of the various models on a held-out test dataset. For these traffic scenarios there are two other useful metrics we can compare: 
the fraction of offroad (i.e. an agent is on the undrivable area), collision (i.e., an agent overlaps with another agent) and total infractions (offroad or collision) the model makes. We note that these metrics are sensitive indicators of model fitness, in the sense that the training data contains no offroad or colliding data examples.  Samples that exhibit these infractions are evidence of model error,
and additionally inform whether modeling mistakes are made globally (i.e. offroad) or through interactions (i.e. collisions). We report our results in \cref{tbl:initial_conditions_baseline}.

\begin{table}[t]
     \caption{Results for scene generation and bounding box prediction.}
     \label{tbl:results}
    \begin{subtable}[t]{0.62\textwidth}
      \caption{
  Quantitative results for traffic scene generation. NLL indicates negative log-likelihood in nats, while the other metrics indicate the fraction of samples exhibiting offroad, collision, or either (lower is better). }
  \label{tbl:initial_conditions_baseline}
        \centering
  \begin{tabular}[b]{lrrrr}
    \toprule
  Method & NLL  & Offroad & Collision & Infraction\\
  \midrule
 Gaussian & 46.2&0.99&0.28&0.99 \\
RealNVP & 32.4&0.97&0.25&0.98 \\
CNF & 22.1&0.88&0.33&0.93 \\
Autoregressive & 13.9&0.74&0.16&0.79 \\
\midrule 
PIF Single & 7.6&{\bf 0.10 } &0.57&0.62 \\
PIF Pair & 6.6&0.19&0.20&0.35 \\
PIF (ours) & {\bf 6.1} &0.12&0.08&{\bf 0.19 } \\
\midrule 
Cond. single & 7.6&0.13&0.12&0.23 \\
Cond. pair & 7.2&0.16&0.15&0.29 \\
Cond. base & 14.0&0.91&{\bf 0.05 } &0.91\\
  \bottomrule
\end{tabular}
    \end{subtable}
    \hfill
    \begin{subtable}[t]{0.35\textwidth}
  \caption{\label{tbl:detection} Quantitative results for bounding box prediction. IOU refers to the intersection over union of object area covered by the samples (higher is better).}    
        \centering
  \begin{tabular}[b]{lr}
    \toprule
    Dataset & IOU \\
      \midrule
      CLEVR3 & 0.732
 \\
      \midrule
      CLEVR6-3 & 0.759 \\
CLEVR6-4 & 0.711 \\
CLEVR6-5 & 0.644 \\
CLEVR6-6 & 0.596 \\
CLEVR6 & 0.679
 \\
  \bottomrule
\end{tabular}
     \end{subtable}
\end{table}

Comparing the likelihood and infraction metrics demonstrates the clear advantage of
using a permutation invariant model. The non-permutation invariant version of our flow does not converge to a competitive likelihood, and struggles to generate infraction-free examples. While the canonically ordered autoregressive model is much more capable than the non-permutation invariant flows, it still underperforms compared to ours. The two ablations of our model
provide an insightful result: the function of $g_{\theta}$ is to define the collective
behavior of the agents (all of them need to stay on the road, independently of
one another), whereas $f_{\theta}$ provides the necessary interaction between them
(agents should not collide with one-another). These functions are evident from
the respective infraction metrics. Furthermore, there appears to be a certain amount of competition between the pair and single terms: as the agents are steered onto the road, they have a higher density and thus a higher chance to collide. The opposite is equally true, as repelling agents can push each other off-road. As such it is not terribly surprising that the ``single-only'' ($g_{\theta}$-based) flow performs better when only considering offroad infractions. Nevertheless, the combined flow has the best performance overall, both in overall infraction rate and negative log likelihood. 

\subsubsection{Conditioning ablations}
To better understand the effects of our conditioning inputs, we report the performance of two ablations, one where the conditional input is only used in $g_{\theta}$  (``Cond. Single''), and one where it is only used in $f_{\theta}$ (``Cond. Pair''). In the former, the interactions between actors are independent of the scene, which one may expect to be a reasonable approximation. However, the results in \cref{tbl:initial_conditions_baseline} indicate that this input is in fact important, which may be understood from the fact that different locations lead to different traffic configurations. In the case where conditioning on $g_{\theta}$ is ignored, we also do worse than conditioning both, while surprisingly maintaining a fairly competitive result. Finally, we show a variant of our model where we condition the base distribution (i.e.~$p_{\theta}(\mathbf{z}|\mathbf{y})$ vs.~$p(\mathbf{z})$), but otherwise remove the conditional dependence from $f_{\theta}$ and $g_{\theta}$. Although in this case the collision rate is lowest, this can be attributed to the poor performance with respect to offroad infractions. Overall, this variation fails to provide a competitive result.
\subsubsection{Variable set size}
Since our model is trained on a variety of different set sizes, and
performs well for each of the different set sizes, we can investigate whether
the it generalizes beyond the number of agents it has seen during
training. We therefore generate samples in our roundabouts model with a previously unseen number of
agents. Such samples are presented in the last column of \cref{fig:initial_conditions_samples}, which have 28 agents, while
the maximum number of agents present in the training data is 22. These results
indicate the the inductive bias of the representation in
\cref{eqn:conditional_dynamics} not only performs well in sample with respect to set size, but
generalizes to larger set sizes too.

\begin{figure*}[t]
  \centering
  \includegraphics[width=5.5in]{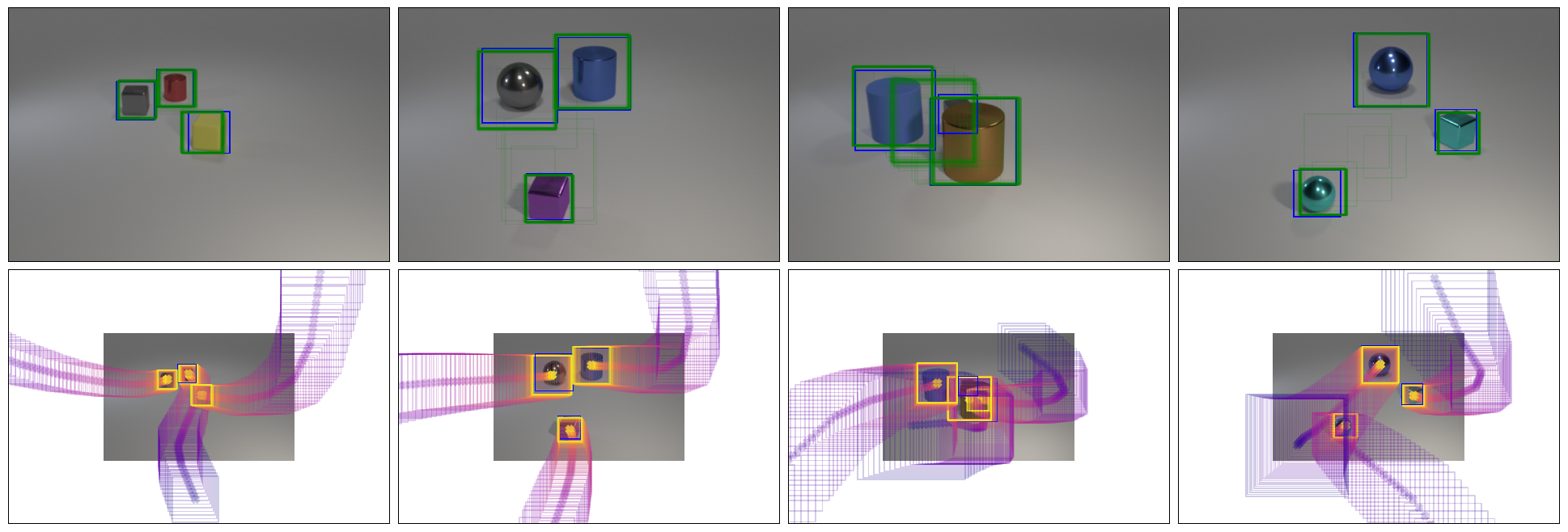}
\caption{\label{fig:clevr3}Bounding box prediction on CLEVR3 images. Each image in the top row shows 50 samples
  from our conditional flow, conditioned on the background image.
  The blue boxes show the ground truth bounding boxes, while the many green boxes
  are conditional flow samples. The bottom row shows the trajectories
  of the boxes for a single sample for each of the same four images. The time
  along the trajectory is encoded with the color of the box. 
  }
\end{figure*}

\subsection{Bounding box prediction}\label{ssec:experiments_detection}
Our final experiment considers object detection, which is a task traditionally divided into bounding box
prediction and classification. The gold standard in this flavor of object
detection remains so-called ``non-max-suppression,'' in which a large number of
putative bounding boxes are scene-conditionally generated and then ``pruned'' by a greedy selection
algorithm~\cite{Girshick2014, Girshick2015}. While the field has shifted focus
toward segmentation in recent years, some very recent studies~\cite{Hu2018,
  Zhang2019, Carion2020} have proposed the use of conditional set generators for object detection. 
We build on this idea by illustrating how our flow can be used to conditionally generate and score sets of object bounding boxes. Being able to compute the log density of a configuration of bounding boxes
eliminates the need for a differentiable matching algorithm~\cite{Hu2018,
  Carion2020}. Having a tractable density also opens up uncertainty-preserving approaches to downstream
tasks such as outlier detection and object counting.

\begin{figure*}[t]
  \centering
  \includegraphics[width=5.5in]{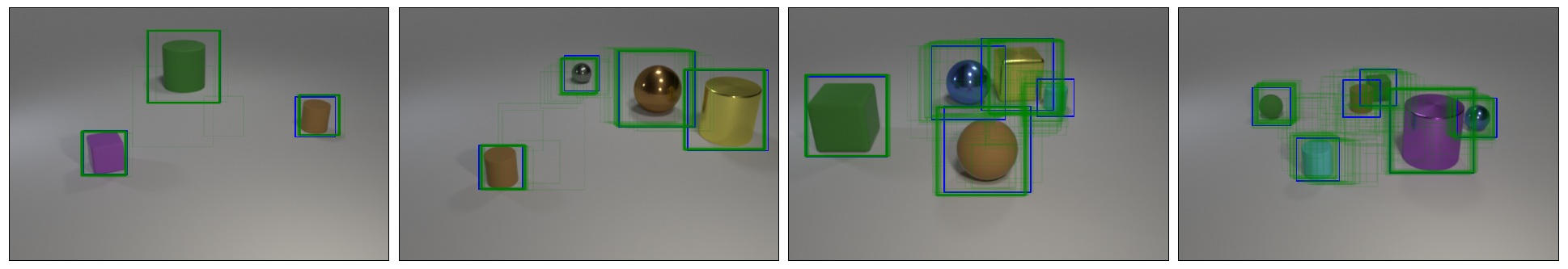}
\caption{\label{fig:clevr6}Bounding box prediction on CLEVR6 images. Each image shows 50 samples
  from our flow conditioned on the corresponding background image.
  The number of objects in these images is three to six, going from left to right.
  The blue boxes show ground truth bounding boxes, while the many green boxes
  are all samples from the learned conditional distribution.}
\end{figure*}

\begin{table}[ht!]
  \centering

\end{table}

We assess the viability of our approach through the CLEVR
\cite{Johnson2017} dataset, which is a standard benchmark for set-generation
models \cite{Greff2019, Zhang2019, Locatello2020}. While this former work combines localization with classification (i.e.,
predicting object \emph{position} and \emph{type}), we focus on the
task of bounding box prediction (i.e predicting object \emph{position}
and \emph{size}). The CLEVR dataset does not provide bounding boxes, so we generate ground truth bounding boxes from object metadata. The full details on how to create bounding boxes for the CLEVR dataset are described in \cref{sup:clevr_size}.

We begin with a subset of the CLEVR dataset only containing three objects. This
allows the flow to learn a transformation that always contains the same number
of objects. We find that our flows perform well on this task, and show example
predictions in \cref{fig:clevr3}. For each conditional image (displayed in the
background), 50 samples from the conditional distribution are shown, graphically illustrating the variance of the conditional density. For each column, the bottom row displays the trajectory taken to generate a single one of these samples.
The trajectories show the interactions between the bounding boxes over time, as
they coordinate through the use of repulsive forces.  Note that by sorting out which of the base
distribution samples goes to which of the objects, the flow solves the assignment
problem along the way. It is
worth pointing out that the third sample has an occluded object, and the
variance of the object position is clearly higher than that for objects where
there is no occlusion, which we take to be evidence that these flows for bounding box prediction should be useful in uncertainty aware downstream applications. Importantly, the variance of the size is not significantly increased,
which is correct behavior for this example. Additional samples are presented in the appendix. The averaged IOU
with the ground truth is reported in \cref{tbl:detection}, and corresponds
approximately to a 15\% mismatch on average in each spatial dimension.
\footnotetext{By column from left to right, the flow samples are: top, top, top, bottom, both. For the last column in particular, our flow is able to produce more vehicles than ever appeared on that map in the training data.}

We continue our exploration with a larger subset of the CLEVR dataset, including
images that have between three and six objects. This subset has also been used in
\cite{Locatello2020} for object detection. We assume that the number of objects
is given, and only predict the bounding box locations and sizes given the
number of bounding boxes and the image. 
Some
example samples are displayed in \cref{fig:clevr6}, with green boxes displaying samples
from the distribution that is conditioned on the image in the background. More
examples are given in \cref{sup:clevr6_additional_results}. The flow
generalizes well over set cardinalities, hinting that some generalizing principles are learned by the flow about these bounding boxes interact, even with different box cardinality.
We moreover see that the more crowded the image becomes, the more spread there
is in the predicted bounding boxes, representing increased uncertainty about
object sizes and positions, also representing more occlusion. The overall IOU,
as well as the IOU's separated by set cardinality are given in
\cref{tbl:detection}. These results show that the flow trained on data with
variable set size performs marginally better on the CLEVR3 subset than a flow trained
only on that data, which we speculate is due to the larger amount of available data. A modest decrease in IOU
can be observed as the sets become larger, resulting in an overall performance
that is slightly lower.

\section{Discussion and conclusions}\label{sec:discussion_and_conclusion}
This work introduced conditional permutation invariant flows, a framework built on
continuous normalizing flows that enables conditional set generation with a
tractable density. We have applied our flows to two problems: realistic traffic scene generation, and bounding box prediction. For traffic scene generation, we significantly outperform baselines, and are the first to present a permutation invariant solution. Ablations to our flows highlight their intuitive mechanism, that can be understood as objects moving jointly in a field, augmented with pairwise interaction potentials. We have moreover shown that bounding box prediction can be enhanced with a tractable log density, potentially opening an avenue to develop downstream vision algorithms that deal with uncertainty in a more principled way.
There are numerous directions for future improvement of the proposed method. Notably, our pair wise interaction comes at a quadratic cost in the number of particles, which complicates extending the method to larger set sizes. We suggest this could be addressed by creating a finite interaction horizon for objects. 

\begin{ack}
We acknowledge the support of the Natural Sciences and Engineering Research Council of Canada (NSERC), the Canada CIFAR AI Chairs Program, and the Intel Parallel Computing Centers program. Additional support was provided by UBC's Composites Research Network (CRN), and Data Science Institute (DSI). This research was enabled in part by technical support and computational resources provided by WestGrid (www.westgrid.ca), Compute Canada (www.computecanada.ca), and Advanced Research Computing at the University of British Columbia (arc.ubc.ca). TC acknowledges support from a National Sciences and Engineering Research Council of Canada (NSERC)
Discovery Grant and a gift from Google LLC.
\end{ack}

\clearpage

\bibliographystyle{abbrv}
\bibliography{equiflows}

\clearpage
\appendix
\section{Appendix}
\subsection{Proof}\label{sup:proof}
We provide here the proof for \cref{thm:perm_invariant}.
\bprf
We will
consider equivariance of $T$ to a transposition of elements $i$ and $j$, denoted
$\sigma_{i,j}$. A transposition $\sigma_{i,j}$ is a permutation for which
$\sigma_{i} = j$, $\sigma_{j} = i$, and $\sigma_{k} = k$ for all $k \in \{1 \dots N\} \setminus \{i,j\}$. Since any permutation can be constructed from a series of transpositions, proving that $T$ is equivariant to a transposition, trivially extends to equivariance to all permutations. In the following we have dropped the dependence on $\mathbf{y}$ and $t$ for notational clarity. We have
\newcommand{\sij}{\sigma_{i,j}} \renewcommand{\bx}{\mathbf{x}}
\renewcommand{\bv}{\mathbf{v}} \renewcommand{\bxi}{\mathbf{x}_{i}}
\newcommand{\bxj}{\mathbf{x}_{j}} \newcommand{\bxk}{\mathbf{x}_{k}}
\newcommand{\bvi}{\mathbf{v}_{\theta, i}} \newcommand{\bvj}{\mathbf{v}_{\theta,j}}
\newcommand{\bvk}{\mathbf{v}_{\theta, k}}
\[
  T(\sigma_{i,j}\mathbf{x}) =\sij\bx +  \int_{t_{0}}^{t_{1}}\mathbf{v}_{\theta}(\sigma_{i,j}\mathbf{x}) dt.
\]
The first term $\sij\bx$ trivially satisfies the equivariance condition.
Focusing on the $i^\text{th}$ term of the dynamics function $\mathbf{v}_{\theta,i}$
\begin{align}
  \bvi(\sij\bx) =& \sum_{k \setminus \{i\}} f_{\theta}\left((\sij \bx)_{i}, (\sij \bx)_{k} \right) + g_{\theta}\left((\sij \bx)_{i} \right) \\
                =& \sum_{k \setminus \{i,j\}} f_{\theta}\left((\sij \bx)_{i}, (\sij \bx)_{k} \right) + f_{\theta}\left((\sij \bx)_{iv}, (\sij \bx)_{j}\right) +  g_{\theta}\left(\bxj \right) \\
                =& \sum_{k \setminus \{i,j\}} f_{\theta}\left(\bxj, \bxk \right) + f_{\theta}\left(\bxj, \bxi\right) +  g_{\theta}\left(\bxj \right) \\
                = & \sum_{k \setminus \{j\}} f_{\theta}\left(\bxj, \bxk \right) +  g_{\theta}\left(\bxj \right) \\
                =&~\bvj \left( \bx \right)  = (\sij \bv_{\theta}(\bx))_{i},
\end{align}
thus demonstrating the dynamics are equivariant and
\[
  T(\sigma_{i,j}\mathbf{x}) = \sij\bx + \int_{t_{0}}^{t_{1}}\sij\mathbf{v}_{\theta}(\bx) dt = \sij T(\bx).
\]
For continuous normalizing flows, the inverse transformation $T^{-1}(\bx)$ is obtained by reversing the integration limits, and an identical derivation can be made to show $T^{-1}(\bx)$ is also equivariant.
For the density, we have
\[
  \log p_{t_{1}}(\sij\bx(t_{1})) = \log p_{t_{0}}\left(T^{-1} \left(\sij \bx(t_{1})\right)\right)
  - \int_{t_{0}}^{t_{1}} \nabla_{\mathbf{x}} \cdot \bv(\sij\bx) dt.
\]
Here, the divergence $\nabla_{\mathbf{x}} \cdot \bv_{\theta}\left(\sij\bx\right)$ denotes the divergence with respect to the argument of $\bv$, evaluated at $\sij \bx$. Since the base distribution $p_{t_{0}}$ is
permutation invariant, and $T^{-1}$ is equivariant
\[
  \log p_{t_{0}}\left(T^{-1} \left(\sij \bx(t_{1})\right)\right)
  &= \log p_{t_{0}}\left(\sij T^{-1} \left(\bx(t_{1})\right)\right) \\
  &= \log p_{t_{0}}\left(T^{-1} \left(\bx(t_{1})\right)\right).
\]
The divergence in the second term $\nabla_{\mathbf{x}} \cdot \bv_{\theta}\left(\sij\bx\right)$ is a sum over derivatives with respect to all its arguments, so is invariant to $\sij$ 
\[
  \nabla_{\mathbf{x}} \cdot \bv_{\theta}\left(\sij\bx\right) = \nabla_{\mathbf{x}} \cdot \bv_{\theta}\left(\bx\right).
\]
We therefore have
\[
  \log p_{t_{1}}(\sij\bx(t_{1})) = \log p_{t_{1}}(\bx(t_{1})),
\]
thus showing that the dynamics are \emph{equivariant}, and the density is \emph{invariant}.
\eprf

\subsection{Experimental details}\label{sup:experimental_details}
\subsubsection{Force Function Implementation}\label{sup:force_functions}
We model the force functions $f_{\theta}(\mathbf{x}_{i},\mathbf{x}_{j})$ and
$g_{\theta}(\mathbf{x}_{i})$ by feed forward neural networks. For the pair force
$f_{\theta}(\mathbf{x}_{i},\mathbf{x}_{j})$, we concatenate the inputs
$\mathbf{x}_{i}$ and $\mathbf{x}_{j}$. In the case a time variable ($t$) is used, it
is also concatenated. For the conditional input, we
construct an embedding vector $\mathbf{y}_{\mathrm{emb}}$, which we concatenate
to the second layer inputs of $g_{\theta}$ and $f_{\theta}$. The embedding vector
$\mathbf{y}_{\mathrm{emb}}$ is generated using a separate neural network, here chosen
to be a convolutional network, since all our conditional distributions have
images as inputs. 
\subsubsection{Solving the ODE}
We use the adaptive solver of Dormand and Prince of order 4 to solve the ODE~\cite{Dormand1980}. To calculate the gradients of the ODE with respect to its parameters we use the adjoint method \cite{Coddington1955}. This enables
calculation of the gradients without back propagating through the computational
graph. This functionality is all available in the \texttt{torchdiffeq} package \cite{Chen2018}, which is the implementation we use in our experiments.

\subsubsection{Table of experimental hyperparameters}\label{sup:exp_params}
In all experiments $g_{\theta}$, $f_{\theta}$ are implemented as neural networks of $n$ layers and $h$ neurons per layer. The convolutional embedding network has $n$ layers of $c$ channels, followed by a single feed forward layer of $h$ neurons. We use sigmoid-linear units in all our dynamics functions, which satisfy the requirement of Lipschitz continuity, provided the networks are evaluated on a finite domain. The use of a time variable in the dynamics is indicated with $t \in \{\text{yes}, \text{no} \}$. For each experiment, these parameters are presented in \cref{tbl:exp_params}.
\begin{table}[t]
     \caption{Hyperparameters used for experiments. Abbreviations are defined in the appendix text.}\label{tbl:exp_params}
     \centering
  \begin{tabular}[t]{lccccccccc}
    \toprule
  Experiment & \multicolumn{2}{c}{$g_{\theta}$}  & \multicolumn{2}{c}{$f_{\theta}$} & \multicolumn{3}{c}{$\mathbf{y}_{emb}$}  &  \\
  \cmidrule(lr){2-3}
  \cmidrule(lr){4-5}
  \cmidrule(lr){6-8}

        &  $n$  &   $h$   &  $n$   &   $h$  &  $n$ & $c$  &  $h$  & $t$ & batch \\
  \midrule
  Example conditional  & 5 & 200 & 5 & 200 & 3 & 16 & 200 & yes & 100 \\
  Example bounding box & 5 & 200 & 5 & 200 & 3 & 16 & 200 & yes & 100 \\ 
  \midrule
  Traffic baseline     & 5 & 200 & 4 & 100 & 3 & 32 & 500 & no  & 100 \\
  \midrule
  CLEVR3               & 4 & 188 & 5 & 196 & 5 & 18 & 409 & no  & 100 \\
  CLEVR6               & 5 & 100 & 5 & 200 & 5 & 28 & 478 & no  & 100 \\
  \bottomrule
\end{tabular}
\end{table}

\subsubsection{Computational Cost}\label{sup:computational_cost}
Summing over all pairs of interactions is necessary to make the transformation
permutation equivariant, but it comes at a quadratic cost in $N$. While not
problematic for the set sizes in this study, this is clearly a limited approach
for large numbers of objects. In these cases, it would be possible to set a
boundary on the interaction range, or use a fixed set of $M < N$ inducing
points, for a total cost of $MN$. Such approximations have been studied for
example in transformers (which are also quadratic in the sequence
length)~\cite{Vaswani2017, Wang2020}. Furthermore, the divergences with respect to
$\mathbf{x}_{i}$ are still quadratic with respect to $D$. This has been
addressed in recent work by using functions that have divergences that can be
easily evaluated using automatic differentiation \cite{Chen2019, Bilos2021}.
Although these types of functions are compatible with our framework, the current
work only considers cases where $D \leq 5$, and therefore we do not implement it. Our overall
algorithm therefore is of cost $N^{2}D^{2}$. If it is necessary to
construct distributions with larger $D$ (in for example object detection, rather
than bounding box prediction), it is possible to use the methodologies from the
aforementioned work to end up with a total computational cost of $MND$.
\subsubsection{Computational resources}\label{sup:computational_resources}
All our experiments were performed on a single GPU, all permutation invariant models were trained between 2 and 7 days of wall-clock time. The ``vanilla'' continuous normalizing flow, realNVP, and autoregressive model were trained over 14 days of wall-clock time.
\subsubsection{Rejection sampling}
All samples presenting traffic scenarios in the main script have been checked by an automated procedure, and in a small amount of cases were rejected if an infraction occurred based on actors being offroad or vehicle overlap.
\subsubsection{Datasets}
The datasets used in this study are the INTERACTION datset \cite{interactiondataset} (available for research purposes), and the CLEVR dataset \cite{Johnson2017} (available under the Creative Commons CC BY 4.0 license).

\subsection{Additional Experimental results}\label{ssec:addtional_experimental_results}
\begin{figure*}[t]
  \centering
  \includegraphics[width=4.0in]{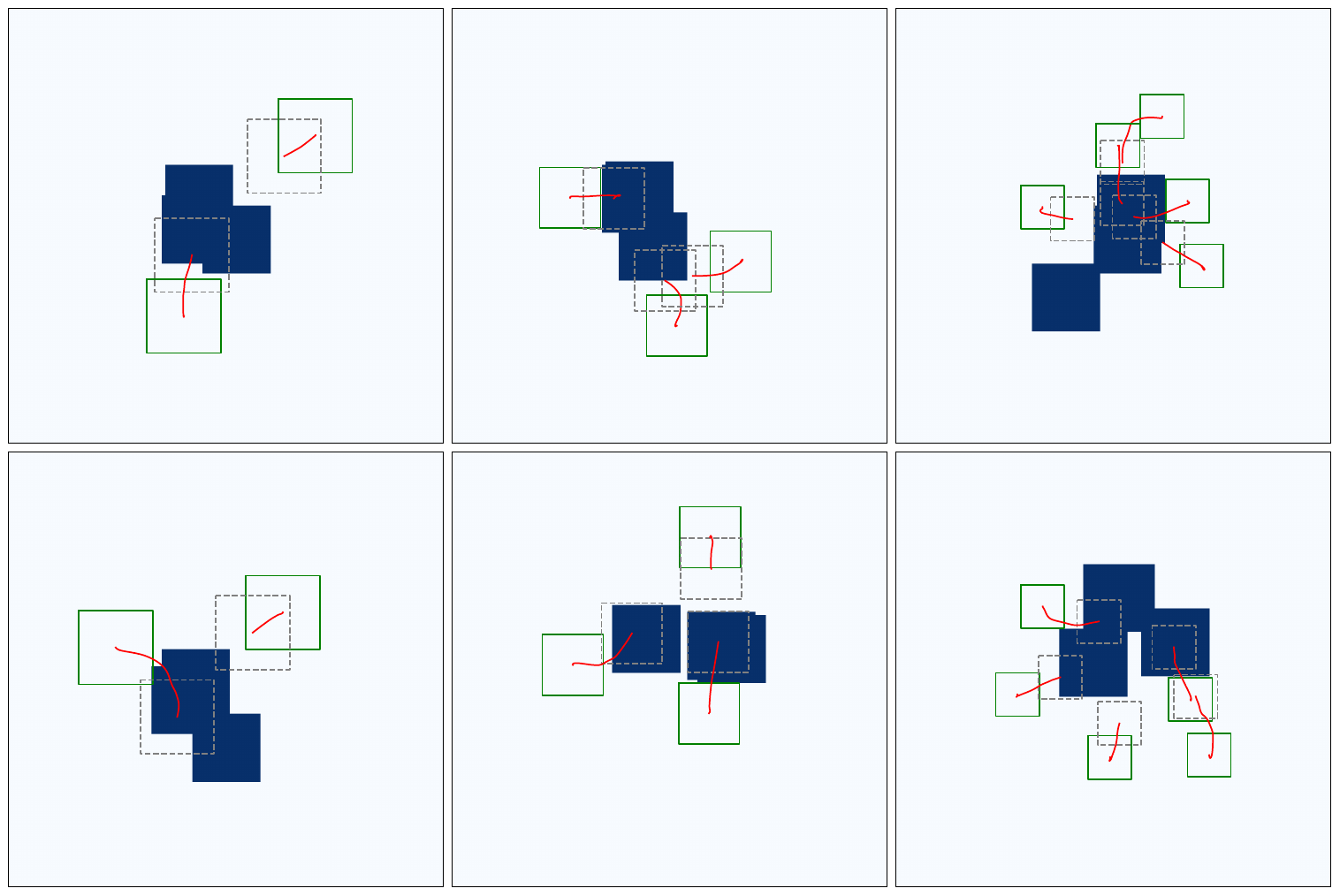}
\caption{\label{fig:cond_sampling}Conditional samples. The condition is an image, which is plotted as a
  blue on white background. The distribution is trained on samples that do not
  overlap with the blue regions, or with oneanother. The grey boxes are samples from the base
  distribution, the green boxes are samples from the flow. The red curves
  indicate the traveled trajectory for each box.}
\end{figure*}
\begin{table}[t]
  \centering
    \caption{\label{tbl:cond_boxes_table}Acceptance rates for conditional sampling. Results presented are the
    acceptance rate (AR) for Prior samples and the Conditional Permutation Invariant Flows (PIF) }
  \begin{tabular}{ccc}
  \toprule
  Set Size & Prior AR & PIF AR \\
\midrule
2 & 0.01 & 0.83 \\
3 & $2.01 \cdot 10^{-3}$  & 0.79 \\
5 & $1.76 \cdot 10^{-4}$ & 0.77
 \\
  \bottomrule
\end{tabular}
\end{table}

\subsubsection{Conditional non-overlapping boxes}
\label{sec:cond_non_overlapping_boxes}
\begin{figure*}[t]
  \centering
  \includegraphics[width=4.5in]{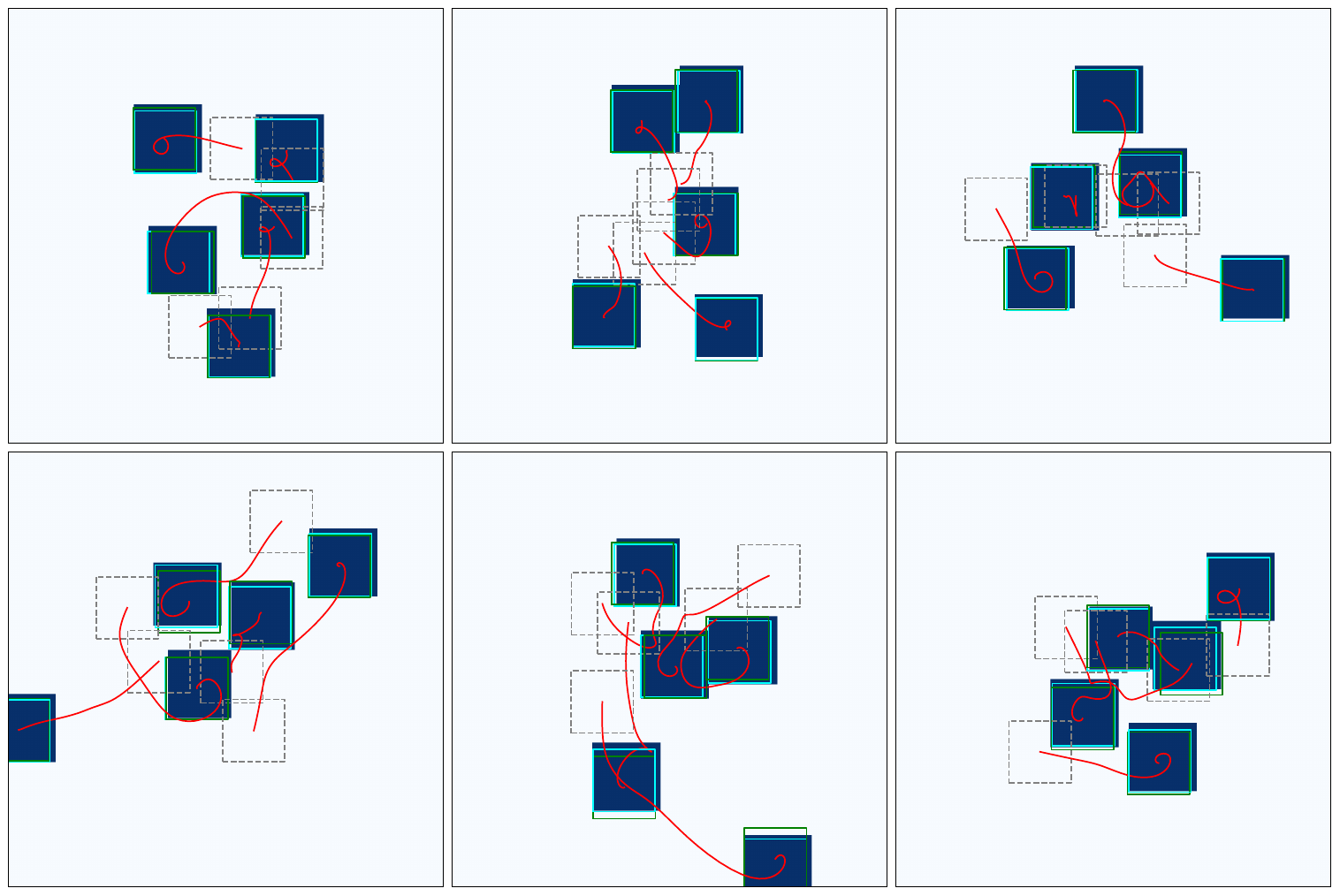}
\caption{\label{fig:toy_detection} Additional samples of the example bounding box prediction problem.}
\end{figure*}

Additional samples of of the conditional generation of non-overlapping boxes (\cref{sec:synthetic}) are presented in \cref{fig:cond_sampling}, for cardinalities 2, 3 and 5. Performance in terms of acceptance rate against an independent prior are reported in \cref{tbl:cond_boxes_table}.

\subsubsection{Bounding box prediction}
\label{sec:toydetection}
Additional samples for the example bounding box prediction task (\cref{sec:synthetic}) are presented in \cref{fig:toy_detection}.

\subsubsection{Outlier detection}
\begin{figure*}[t]
  \centering
  \includegraphics[width=4.5in]{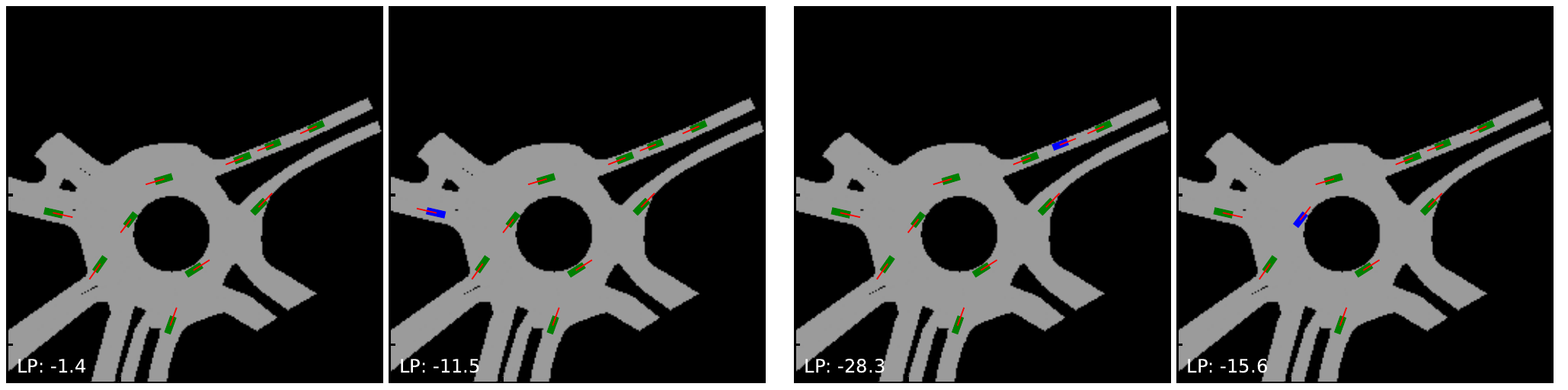}
\caption{\label{fig:outlier_detection}
Samples from the distribution with more agents than occur in the dataset (first
two panels). Sample and its log probability (in \emph{nats}) and 2 corrupted variations
where actors (in blue) have been turned around (last three panels).}
\end{figure*}

Since our model has a tractable density, we can use it for outlier detection. In the traffic scene task, we study the case of mislabelled examples, which we artificially generate
by rotating one of the actors in the scene by $\pi$. The original and corrupted
scenes and corresponding log probabilities are shown in the last three panels of
\cref{fig:outlier_detection}. It is
clear that reversing one of the actors substantially decreases the probability. Moreover
the model correctly captures the severity of the resulting
infraction, which is less when an actor is going the wrong way on a two-way road, without the presence of other surrounding actors.

\subsubsection{Regularization}\label{sup:regularization}
\begin{figure*}[t]
  \centering
  \includegraphics[width=5.5in]{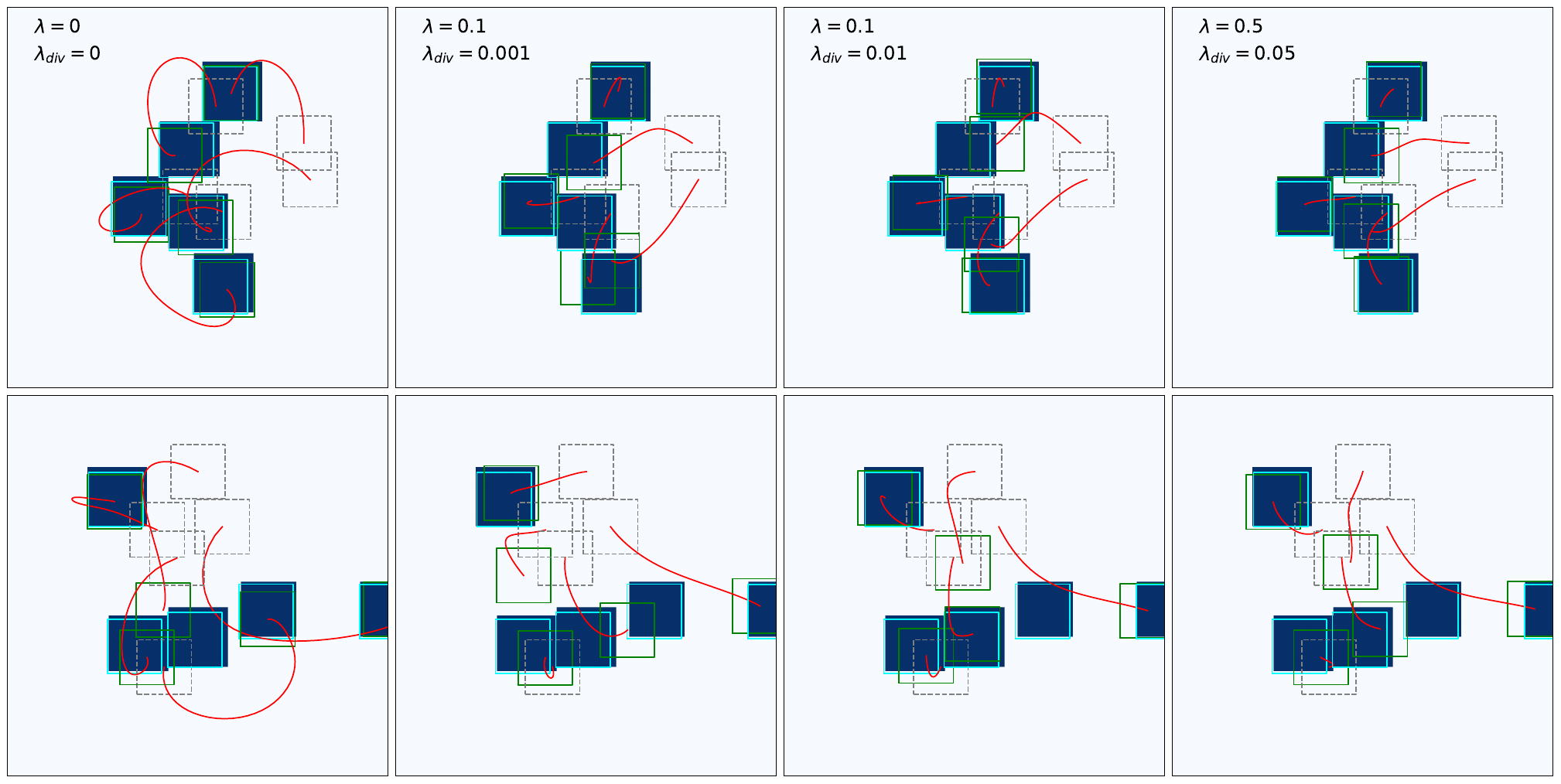}
\caption{\label{fig:regularization} The effect of regularization on the dynamics. The penalty proportionality constants are reported per column in the top panel.}
\end{figure*}

We present the effect of our regularization term on the bounding box prediction task, in which the effect is most pronounced. The proportionality constants of the $\ell^2$ and  $\ell^2_{div}$ penalty terms are denoted as $\lambda$ and $\lambda_{div}$ respectively. Results for various $\lambda$ and $\lambda_div$ are presented in \cref{fig:regularization}. The increase of the parameters $\lambda$ and $\lambda_div$ evidently creates more direct trajectories. We further find empirically that it drastically reduces the number of calls to the dynamics made by the ODE solver.

\subsubsection{CLEVR3}
\begin{figure*}[t]
  \centering
  \includegraphics[width=5.5in]{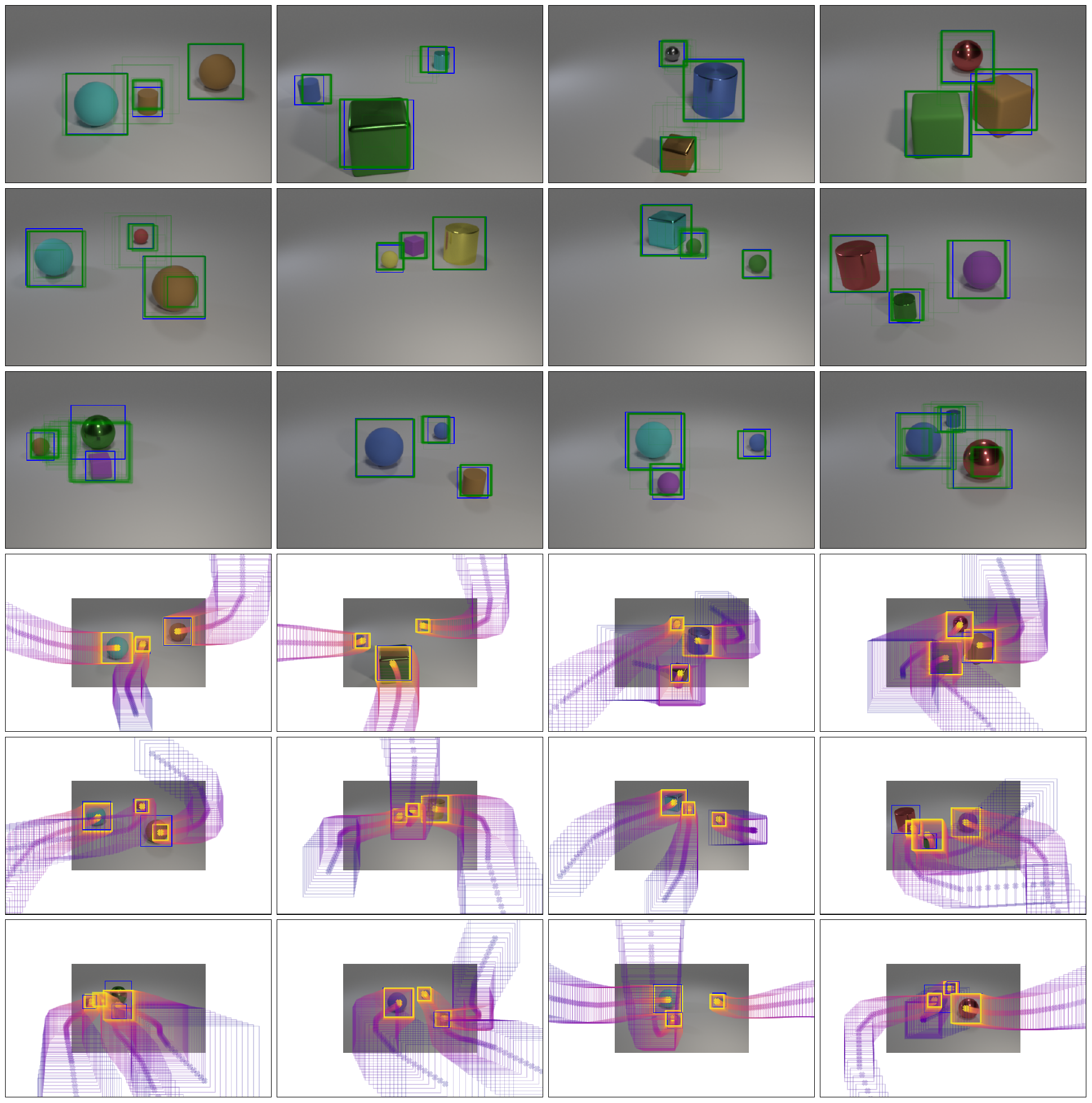}
\caption{\label{fig:clevr3_sup} Additional examples for the CLEVR3 dataset}
\end{figure*}

Additional samples conform presentation in the main text are presented in \cref{fig:clevr3_sup}.
\label{sup:clevr3_additional_results}

\subsubsection{CLEVR6}
\begin{figure*}[t]
  \centering
  \includegraphics[width=5.5in]{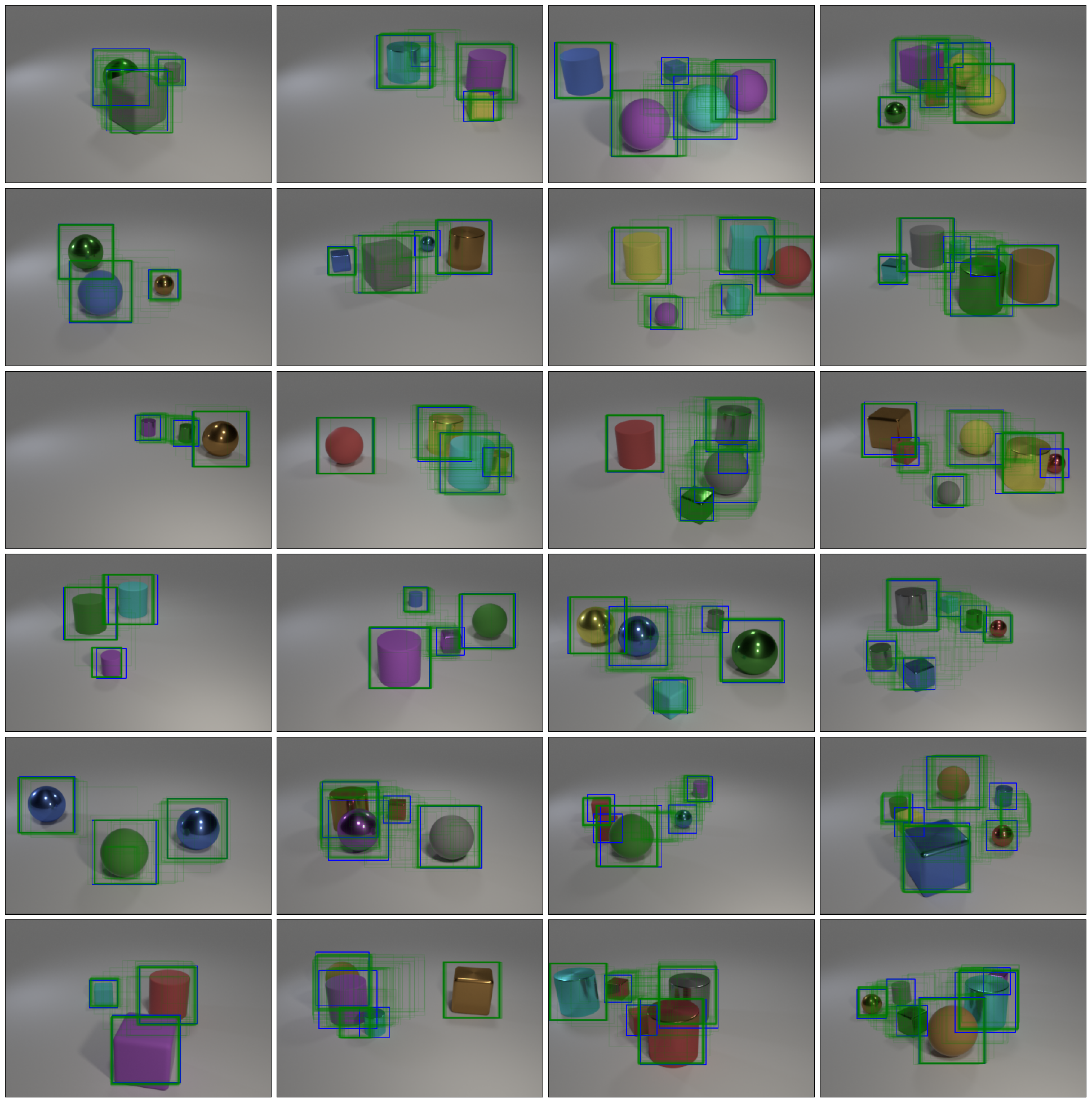}
\caption{\label{fig:clevr6_sup} Additional examples for the CLEVR6 dataset}
\end{figure*}

Additional samples conform presentation in the main text are presented in \cref{fig:clevr6_sup}.
\label{sup:clevr6_additional_results}

\subsection{CLEVR size}
\label{sup:clevr_size}
The CLEVR dataset contains the pixel positions of objects, but not the bounding
box sizes. The dataset does however provide the center locations of the objects
in the global frame $\{x_{i},y_{i},z_{i}\}$. Since the objects are sitting on a
flat plane at $z = 0$, its $z$ coordinate is equal to half its size.
Furthermore, the dataset provides the distance to the camera along the viewpoint
axis, $d_{z}$. Using these quantities, we approximate the bounding box size
$\Delta$ as:
\[
  \Delta \approx \frac{z_{i}}{\sqrt{d_{z}}}.
\]
We find empirically that this results in reasonable bounding boxes.

\end{document}